\documentclass{article} % For LaTeX2e
\usepackage{iclr2025_conference,times}

\usepackage{amsmath,amsfonts,bm}

\def\eqref#1{equation~\ref{#1}}
\def\1{\bm{1}}

\DeclareMathAlphabet{\mathsfit}{\encodingdefault}{\sfdefault}{m}{sl}
\SetMathAlphabet{\mathsfit}{bold}{\encodingdefault}{\sfdefault}{bx}{n}

\usepackage{hyperref}
\usepackage{url}

\usepackage{wrapfig}

\usepackage{multirow}
\usepackage{amsmath}
\usepackage{amssymb}
\usepackage{booktabs}
\usepackage{array} % 使用 array 宏包
\usepackage{makecell} % 使用 makecell 宏包
\usepackage{wrapfig} % 加载 wrapfig 宏包
\usepackage{colortbl}
\usepackage{graphicx}

\title{EMMA: Empowering Multi-modal Mamba with Structural and Hierarchical Alignment}

\author{Yifei Xing\textsuperscript{1,2}, Xiangyuan Lan\textsuperscript{1,*}, Ruiping Wang\textsuperscript{2,3,*}, Dongmei Jiang\textsuperscript{1}, Wenjun Huang\textsuperscript{4}\\ \textbf{Qingfang Zheng\textsuperscript{1}, Yaowei Wang\textsuperscript{1}}\\
\textsuperscript{1}Peng Cheng Laboratory, Shenzhen, 518000, China, \\
\textsuperscript{2}University of Chinese Academy of Sciences, Beijing, 100049, China, \\
\textsuperscript{3}Institute of Computing Technology, Chinese Academy of Sciences, Beijing, 100190, China, \\
\textsuperscript{4}Sun Yat-sen University, Guangzhou, 510275, China \\
\texttt{\{xingyf, lanxy, jiangdm, zhengqf01, wangyw\}@pcl.ac.cn} \, \\
\texttt{wangruiping@ict.ac.cn, huangwj98@mail2.sysu.edu.cn} 
}

\begin{document}

\maketitle

% Mamba模型缺乏对图像细粒度和结构的学习，概括现象的根本原因

% 搜一下结构化表征，卷集合多尺度，spatial structure，swin transformer

% are more computationally efficient while showing potential 

\begin{abstract}
Mamba-based architectures have shown to be a promising new direction for deep learning models owing to their competitive performance and sub-quadratic deployment speed. However, current Mamba multi-modal large language models (MLLM) are insufficient in extracting visual features, leading to imbalanced cross-modal alignment between visual and textural latents, negatively impacting performance on multi-modal tasks. In this work, we propose \textbf{E}mpowering \textbf{M}ulti-modal \textbf{M}amba with Structural and Hierarchical \textbf{A}lignment (EMMA), which enables the MLLM to extract fine-grained visual information. Specifically, we propose a pixel-wise alignment module to autoregressively optimize the learning and processing of spatial image-level features along with textual tokens, enabling structural alignment at the image level. In addition, to prevent the degradation of visual information during the cross-model alignment process, we propose a multi-scale feature fusion (MFF) module to combine multi-scale visual features from intermediate layers, enabling hierarchical alignment at the feature level. Extensive experiments are conducted across a variety of multi-modal benchmarks. Our model shows lower latency than other Mamba-based MLLMs and is nearly four times faster than transformer-based MLLMs of similar scale during inference. Due to better cross-modal alignment, our model exhibits lower degrees of hallucination and enhanced sensitivity to visual details, which manifests in superior performance across diverse multi-modal benchmarks. Code will be provided.

\end{abstract}
\renewcommand{\thefootnote}{}
\footnotetext{*Corresponding authors Xiangyuan Lan and Ruiping Wang}
\section{Introduction}

Recently there has been a notable increase in the development of domain-general AI agents \cite{kalla2023study, zhao2023survey} which can simultaneously solve a diverse range of tasks and exhibit superior performance. Among them, multi-modal large language models (MLLMs) \cite{achiam2023gpt, team2023gemini, liu2024visual} have emerged as a promising direction due to their effectiveness in visual perception and logical reasoning. MLLMs usually consist of an image encoder that converts images to visual tokens and a strong large language model (LLM) backbone to process the visual and textual tokens concurrently. This integration of visual and textual information not only enhances the understanding of visual content but also provides a more comprehensive context for language understanding and generation. As a result, these cross-modal models have consistently achieved state-of-the-art performances in tasks such as image captioning \cite{hossain2019comprehensive}, visual reasoning \cite{johnson2017clevr}, and visual question answering \cite{antol2015vqa}. 

However, a gravid challenge for current MLLMs is the substantial computational cost associated with training and deployment \cite{achiam2023gpt, han2022survey}. In fact, current MLLMs are predominately transformer-based, and consequently suffer from an input-dependent attention mechanism that is quadratic in complexity \cite{katharopoulos2020transformers}. Transformers also struggle with capturing long-ranged dependencies in data due to a limited context window \cite{zimerman2023long, tay2020long}. Many works have attempted to address these issues and propose novel architectures that increase effectiveness in long-sequence processing while reducing both computational and memory costs \cite{wang2020linformer, shen2021efficient, li2020linear, gu2023Mamba}. 

Mamba \cite{gu2023Mamba, dao2024transformers} is a structured state-space sequence model (SSM) proposed as an alternative for transformers \cite{aoki2013state, gu2021efficiently, gu2021combining} and has demonstrated superior performance in sequential tasks such as language and genomics \cite{ma2024u, gu2023Mamba}. Inspired by classical state-space models \cite{kalman1960new}, Mamba is efficiently formulated as either a recurrence or convolution with near-linear scaling in sequence length, while also possessing principled mechanisms for modeling long-ranged dependencies. Furthermore, \cite{dao2024transformers, ali2024hidden} have addressed the close connection between Mamba and variants of attention mechanisms in transformers, suggesting a promising new underlying architecture for previously transformer-dominated models that is more efficient and achieves competitive performances. 

However, while Mamba models possess superior performance on sequential textual data, they remain ineffective in processing visual data, especially in large and huge models \cite{ren2024autoregressive, zhu2024vision, liu2024vMamba} where performance degrades quickly when in the billion-parameter scale. This causes an imbalance in the quality of visual and texture latents in the Mamba LLM, where coarse visual features are ineffectively aligned with better-quality textual features. Furthermore, while transformer-based LLM models contain positional embeddings to guide the learning of spatial information, Mamba LLM layers lack structural constraints on the visual features, causing deeper layers of the LLM to gradually lose fine-grained spatial integrity in visual features, as shown in Fig. \ref{fig:compare}. This eventually leads to an increasingly blurred feature during the final layers of the LLM, which weakens the cross-modal alignment between visual and textual cues, ultimately resulting in suboptimal performance and higher degrees of hallucination. With respect to improving the quality of visual latents, current Mamba-based MLLMs employ various strategies such as incorporating multiple vision encoders \cite{zhao2024cobra} or integrating a visual selective scan within the projection MLP \cite{qiao2024vl} to impose structural constraints. However, these approaches primarily focus on preserving the visual feature \textit{prior} to it's integration into the Mamba LLM, lacking mechanisms to maintain its quality \textit{within} the MLLM, where the visual latents participates in the autoregressive generation of textual tokens.

% unlike transformer-based models, current Mamba-based MLLMs \cite{qiao2024vl, zhao2024cobra} suffer from two major drawbacks during optimization: 1. They produce coarse visual features that are ineffectively aligned with better-quality textual data. While Mamba LLMs has been shown to be effective in NLP tasks \cite{gu2023Mamba, dao2024transformers}, visual Mamba models \cite{zhu2024vision, liu2024vMamba} experience a performance plateau when scaled up in parameters \cite{ren2024autoregressive}, leading to poor-quality visual features despite correspondingly effective textual features. 2. They tend to lose fine-grained visual cues as the depth of the LLM layers increases. While transformer-based LLM models contain positional embeddings to guide the learning of spatial information, Mamba LLM layers lack structural constraints on the visual features, causing deeper layers of the LLM to gradually lose fine-grained spatial integrity in visual features. These issues culminate in a diminished visual supervisory signal that manifests in a weakened cross-modal alignment between visual and textual cues, leading to suboptimal performance.

\def\figCompare#1{
\begin{figure*}[#1]
    \includegraphics[width=\linewidth]{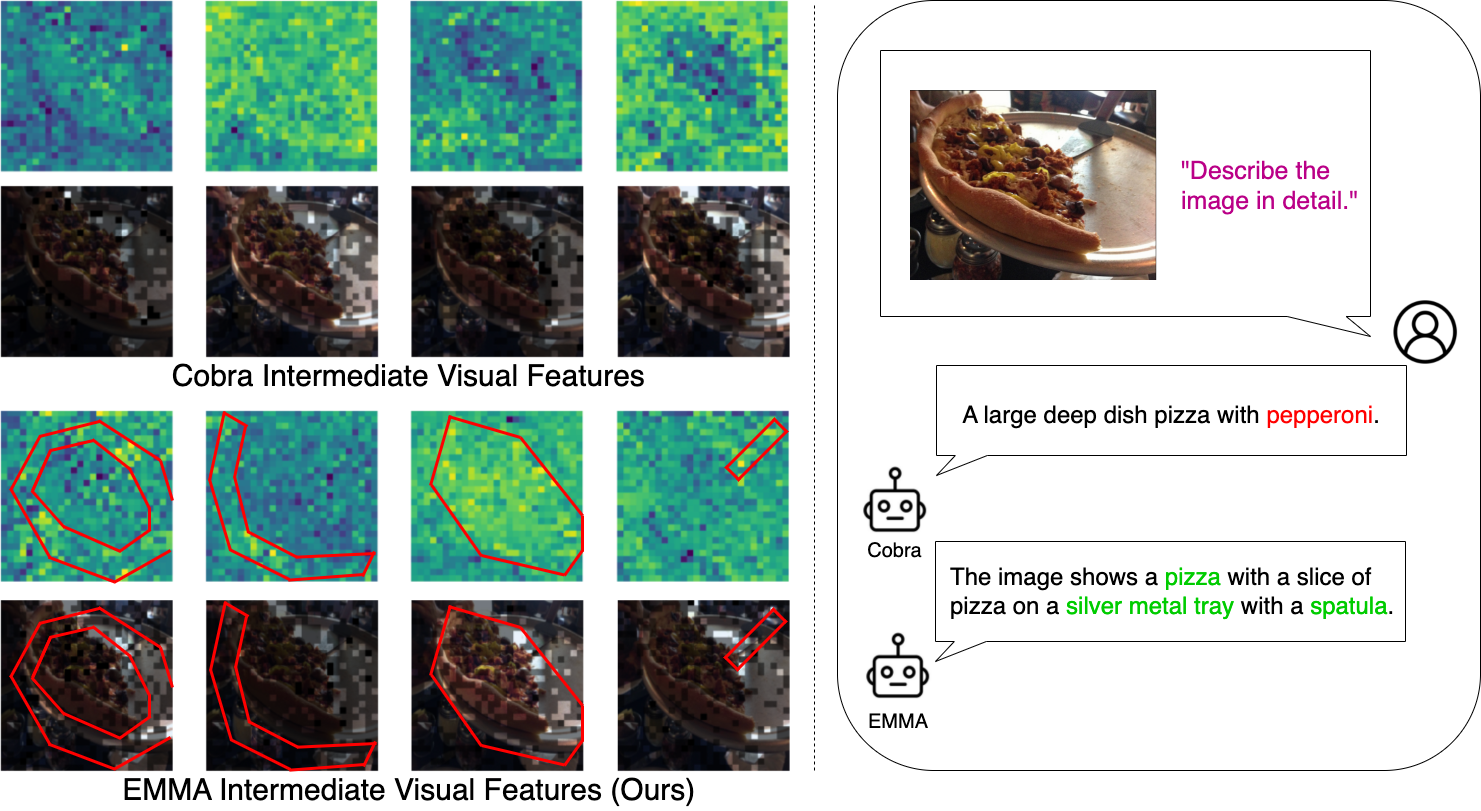}
    \caption{Given an image of a pizza and the prompt 'Describe the image in detail', we visualize intermediate visual features and their corresponding textual responses in Mamba-based MLLMs. Each upper row represents the magnitude of reconstructed spatial activations on each image, and each bottom highlights these patches on the original image. Each column (from left to right) represents intermediate layers with increasing depth. \textbf{Cobra} \cite{zhao2024cobra} experiences a gradual loss of visual features as visual cues become blurred and unrecognizable, resulting in ineffective cross-modal alignment and producing an overly generalized (and hallucinated) answer. On the other hand, due to better cross-modal alignment, \textbf{EMMA} is capable of preserving visual details even in deeper layers of the LLM, highlighting areas such as the perimeter of the pizza tray, the overall pizza, and the spatula on the top right of the image. The resulting text demonstrates higher alignment to the image data in the form of sensitivity to visual details and spatial relationships. }
    \label{fig:compare}
\end{figure*}
}

\figCompare{!t}

Building on these observations, we propose \textbf{E}mpowering \textbf{M}ulti-modal \textbf{M}amba with Structural and Hierarchical \textbf{A}lignment (EMMA) that addresses the insufficient extraction of visual information \textit{inside} the Mamba LLM. First, we note the autoregressive nature of Mamba models that allows for an analogous structure for next-token prediction in the visual domain, which has been shown to be effective in vision-related tasks \cite{ren2024autoregressive, el2024scalable}. Thus, we extend the prediction of text tokens to the prediction of the visual image as well, to serve as a \textbf{structural} constraint on the visual latents for the Mamba LLM. A pixel-wise alignment is proposed to enforce the preservation of key spatial and structural information during training. Next, to mitigate the gradual loss of fine-grained visual features in intermediate LLM layers, we propose a Multi-scale Feature Fusion (MFF) module that \textbf{hierarchically} preserves fine-grained visual features. By building additional contribution of intermediate layers in the final pixel-wise alignment loss, the MFF effectively alleviates gradual information loss, resulting in finer-grained visual features. The major contributions of this paper are three-fold:

% 视觉Mamba不理想，无法有效的scale-up，且规模一大性能就不好，所以制作纯Mamba的mllm比较困难。现在的多模态Mamba也没有针对图片数据进行处理，没有针对图片数据的损失。

% 没有特定的处理视觉部分信息的

% 视觉Mamba不理想，无法有效的scale-up，且规模一大性能就不好，所以制作纯Mamba的mllm比较困难。现在的多模态Mamba也没有针对图片数据进行处理，没有针对图片数据的损失。

% text + image is long-ranged.

% Different from existing that either rely on text-only, or require lots tons of image and text data supervision, our Mamba-based utilizes a Mamba-based decoder to reconstruct

\begin{itemize}
    \item We observe the imbalance between the quality of image and text latents within MLLM Mamba models and propose a pixel-wise alignment to autoregressively encourage the learning and processing of structural visual features, ultimately enabling better cross-modal alignment of visual and textural latents.
    \item To mitigate the gradual loss of fine-grained visual cues in the LLM, we propose a Mamba-based multi-scale fusion module that combines visual features from multiple intermediate layers, resulting in a richer representation for the final alignment phase.
    \item We conduct comprehensive experiments on various multi-modal benchmarks, comparing against current state-of-the-art Mamba-based and transformer-based models. Due to better cross-modal alignment, EMMA achieves competitive performance to similar-scaled ViT and outperforms other Mamba-based models on a variety of tasks while achieving high inference speed.
\end{itemize}

\section{Related Works}

\subsection{Mamba-based Models}
Mamba \cite{dao2024transformers, gu2023Mamba} is a form of state-space model that originally stemmed from classical signal processing theory \cite{kalman1960new}. These sequential models inherently excel at capturing long-range dependencies and are designed to be computationally efficient \cite{gu2020hippo, gu2021combining, gu2021efficiently, gupta2022diagonal}. Expanding upon these works, Mamba \cite{gu2023Mamba} introduces a more effective selection mechanism that is input-dependent and a hardware-aware framework that is computationally efficient. Since then, the Mamba architecture has been successfully applied towards a variety of sequential tasks including NLP \cite{yuan2024reMamba, behrouz2024Mambamixer}, speech \cite{zhang2024Mamba, chen2024rawbMamba}, and motion \cite{wang2024text, zhang2024motion}, rivaling the performance of transformer-based models while being more computationally efficient \cite{qu2024survey}. Mamba has also been adapted to non-sequential data such as vision and point clouds \cite{xu2024visual, yang2024vivim, xing2024segMamba, han2024Mamba3d}, which differs by not adhering to any particular ordering \cite{huang2011learning}. These tasks are more challenging due to the lack of suitable and appropriate structural constraints in Mamba layers to preserve order-invariant information, such as position encoding in their transformer counterparts \cite{vaswani2017attention}. Visual State Space model \cite{liu2024vMamba} introduces a novel 2D Selective Scan (SS2D) module that gathers contextual visual information from various perspectives. Vision Mamba \cite{zhu2024vision} applies Mamba to the Vision Transformer architecture and proposes a bi-directional SSM to better process visual features. However, despite gaining success in small-scale scenarios with less than 100M parameters, these vision Mamba models often falter when scaled-up in more complex scenarios. \cite{ren2024autoregressive} enhances Mamba’s visual capability through autoregressive pretraining, which results in higher classification accuracy over its supervised-trained counterparts and successfully scales visual Mamba models to large and even huge model sizes. In this work, we explore ways to apply the Mamba architecture to multi-modal learning tasks, where large-scale processing of visual features in Mamba layers is required for aligning visual and language modalities.

\subsection{Multi-modal Large Language Models (MLLMs)}
MLLMs take advantage of a powerful large language model (LLM) \cite{chang2024survey, achiam2023gpt} to simultaneously process data from both visual and textual modalities. \cite{cho2021unifying, wang2022image} initially proposed a unified transformer architecture for joint visual and language reasoning tasks that demonstrated better generalizability than conventional vision models. Since then, research on MLLMs have established strong baselines for general-AI assistants \cite{liu2024visual, achiam2023gpt, alayrac2022flamingo}, which typically consists of a vision encoder, a cross-modal projector, and a LLM backbone. While most current MLLMs utilize a text-only loss function for optimization \cite{li2024llava, zhu2024llava, lin2024moe}, EMU \cite{sun2023emu, sun2024generative} takes interleaved visual and textual inputs in a unified training object of predicting next text or image token in an autoregressive way. However, they require multiple training stages and additional computational cost of training a Stable Diffusion \cite{rombach2022high} visual decoder. 

A major drawback of current MLLMs resides in the inherent computational overhead \cite{katharopoulos2020transformers}, predominantly attributed to their foundation in transformer architectures. \cite{ma2024ee, ning2024inf} proposes architectural changes to enable more efficient processing of data in MLLMs. Transformer-based MLLMs also struggle with learning long-range dependencies \cite{zhu2021long, gu2023Mamba}, which is crucial for vision-language tasks. \cite{chen2024dolphin} repurposes the image embedding projector to encode long textual context, improving the MLLM's ability to handle longer sequences. Nevertheless, current progress in improving the computational complexity and learning long-range dependencies mostly depends on architectural changes in the transformer backbone. In this work, we explore how to effectively integrate the Mamba model into the transformer-dominated MLLM domain. Mamba offers a three to five times higher throughput than its transformer counterparts, while capable of processing million-length sequential data.

\subsection{Mamba-based MLLMs}
Mamba has been extended to MLLMs due to their capabilities in both vision and NLP. Current multi-modal Mamba models \cite{qiao2024vl, zhao2024cobra} follow a simple LLaVA-based regimen \cite{liu2024visual} through a pretrained image encoder, Mamba LLM backbone, and a llama-like textual loss. These models utilize efficient down-sampling techniques \cite{chu2024mobilevlm, zhao2024cobra} and multi-direction scanning mechanisms \cite{liu2024vMamba, qiao2024vl} as cross-modal projection modules to align visual and textual information. However, they lack explicit visual supervision during training. Given that Mamba models are inherently less effective in processing visual information in large-scale settings \cite{ren2024autoregressive}, this results in poor-quality visual feature that weakens cross-modal alignment in Mamba MLLMs that easily surpass billions of parameters. Our work seeks ways to address the current bottleneck in mamba-based MLLMs where the mamba LLM extracts insufficient visual feature details that results in disparity in the alignment between images and text features. To this end, we impose additional structural constraints on visual features generated by the Mamba LLM, effectively enhancing the overall efficacy of multi-modal representations and cross-modal alignment.

\section{Method}
% In this section, we present the overall framework for our methodology, as shown in Fig. \ref{fig:overall}. We utilize cobra \cite{zhao2024cobra} as our base architecture, which consists of a joint vision encoder, a MLP projector, and a backbone LLM. Expanding upon this foundation, 

In this section, we present the overall framework for EMMA. EMMA aims to empower the insufficient extraction of visual information in Mamba MLLMs through \textbf{structural} and \textbf{hierarchical} alignment. We first provide the model architecture in Sec. \ref{sec:arch}. Then, we describe our pixel-level visual alignment for preserving structural visual features in Sec. \ref{sec:pixel}, and multi-scale fusion module that hierarchically constrains intermediate features for retaining fine-grained visual cues in Sec. \ref{sec:fusion}.

\def\figOverall#1{
\begin{figure*}[#1]
    \centering
    \includegraphics[width=\linewidth]{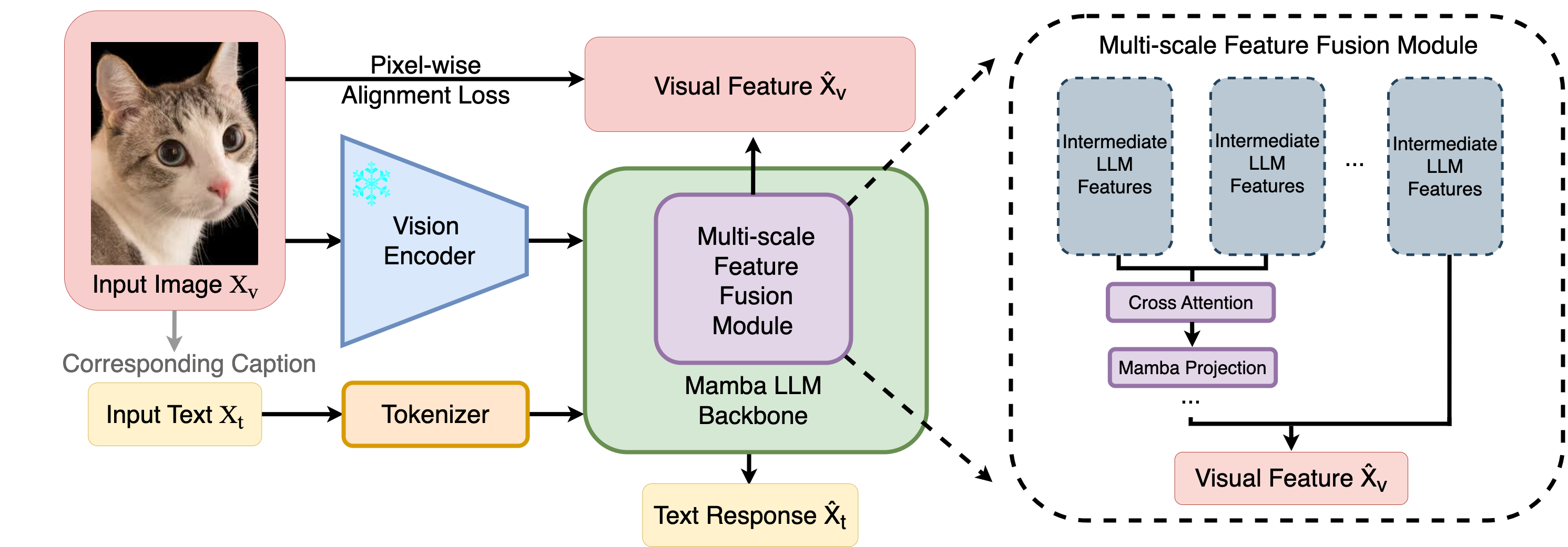}
    \caption{Overview of EMMA. In addition to the textual responses, our model extracts a holistic visual feature from the Mamba LLM through a multi-scale feature fusion module that hierarchically combines intermediate visual features through cross-attention and Mamba projections. A pixel-wise alignment loss is calculated between the final visual feature and the original image to enable the learning of more fine-grained features and increased multi-modal alignment.
    }
    \label{fig:overall}
\end{figure*}
}

\figOverall{!t}

\subsection{Architecture}\label{sec:arch}

Inspired by \cite{zhao2024cobra, qiao2024vl, liu2024visual}, the architecture of EMMA consists of a pre-trained vision encoder $f_v$, a projection MLP $M_{proj}$, a pre-trained Mamba LLM $f_{\phi}$ parameterized by $\phi$, and a multi-scale fusion module $\psi$, as shown in Fig. \ref{fig:overall}. Taking an image $X_v$ and its corresponding tokenized caption $X_t$ as input, we first obtain visual features through the visual encoder: 
\begin{equation}
\Tilde{X_v}=f_v(X_v)
\end{equation}
The MLP then maps visual tokens into the word embedding space, which is concatenated with the tokenized captions to form a multi-modal input. The combined tokens are fed into the Mamba LLM:

\begin{equation}
\Tilde{X}_{LLM}=concat(\Tilde{X_v}, X_t), \quad \hat{X} = f_{\phi}(\Tilde{X}_{LLM}, \psi)
\end{equation}
Where the multimodal response $\hat{X}$ can be separated into a textual response $\hat{X}_t$ and a visual feature $\hat{X}_v$. Note that the visual feature is generated through the multi-scale fusion module which combines intermediate LLM features to alleviate the gradual loss of fine-grained visual features. Lastly, a pixel-wise alignment loss is applied to the visual feature $\hat{X}_v$ for preserving structural visual cues, and an autoregressive NLP loss is applied to the textural feature $\hat{X}_t$. More implementation details can be found in Appendix.

% \subsection{Preliminaries}\label{sec:prelim}

% Recently, \cite{dao2024transformers}

\subsection{Structural Alignment via Pixel-wise Alignment Loss}\label{sec:pixel}

% To incorporate structural supervision on visual features in the Mamba MLLM, we make use of the autoregressively generated visual tokens from the final LLM output. 

We first consider the standard Mamba MLLM training procedure, given multimodal input $\Tilde{X}_{LLM}$ and Mamba LLM $f_{\phi}$:
\begin{equation}
f_{\phi}(\Tilde{X}_{LLM})=[\hat{X}_v, \hat{X}_t]=\hat{X}
\end{equation}
The target text token sequence $\hat{X}_{t} = \{\hat{x}_{t, i}\}^L_i$ of length $L$ is generated by computing the following probability and optimized by minimizing the corresponding negative log-likelihood function $\mathcal{L}_{text}$:
\begin{equation}
p\left(\hat{X}_{t} \mid X_v, X_t \right)=\prod_{i=1}^L p_{\phi}\left(\hat{x}_{t, i} \mid X_v, \{X_{t,j} \mid j < i\} \right), \quad \mathcal{L}_{text}=-\log p\left(\hat{X}_{t} \mid X_v, X_t \right)
\end{equation}
Where $\hat{x}_{t, i}$ depends on the visual features $X_{v}$ and the set of previous text tokens $\{X_{t,j} \mid j < i\}$. We observe the lack of supervision on image-level features, as they serve solely as conditional prior for predicting text tokens $\hat{X}_{t}$. As Mamba models inherently struggle with visual tasks in larger models \cite{ren2024autoregressive}, this may result in poor quality image features during cross-modal alignment, negatively impacting model performance. It becomes crucial to construct additional constraints on the image-level features for better cross-modal alignment. Our insight is to condition the generation of visual features \textit{structurally}, which concern the organization, arrangement, or spatial relationships present within visual data. To achieve this, we extend the paradigm of next-token prediction to enable the joint generation of textural tokens and visual images. Thus, similar formulation arises for the generation of a target visual sequence $\hat{X}_{v} = \{\hat{x}_{v, i}\}^{K}_i$ of length $K$, given by the following probability function:
\begin{equation}p\left(\hat{X}_{v} \mid X_v, X_t \right)=\prod_{i=1}^K p_{\phi}\left(\hat{x}_{v, i} \mid \{X_{v,j} \mid j < i\}, X_t \right)
\end{equation}

% Thankfully, given the unidirectional recurrent structure of Mamba layers, this paradigm of next-token prediction can naturally be extended to generating visual images.

We then project the generated visual feature $\hat{X}_{v}$ from word embedding space back to the image domain through a Mamba-based decoder $f_{dec}$ to retrieve the final generated visual sequence. Using the original image as reference, a common loss function for measuring the quality and consistency between two images is $L2$ distance. The $L2$ loss considers global features of the image, focusing on overall similarity that helps in preserving structures and shapes in the image. We formulate our $\textbf{Pixel-wise Alignment Loss}$ $\mathcal{L}_{pixel}$ to constrain the decoded image to match the original image as:

\begin{equation} \quad \mathcal{L}_{pixel}=\| f_{dec}(\hat{X}_v) - X_v \|^2_{2} 
\end{equation}

By imposing our pixel-wise alignment loss, we constrain the output visual feature to match closely with the original image, forcing the model to understand and preserve structural features of the visual input, and effectively increasing the quality of visual latents for cross-modal alignment. Unlike \cite{zhan2024anygpt, sun2023emu}, our visual decoder is extremely small-scaled and efficient, and does not require a separate stage for training. 

\subsection{Hierarchical Alignment via Multi-scale fusion module}\label{sec:fusion}

While the pixel-wise alignment loss forces the Mamba LLM to retain structural characteristics of the visual input, we find that Mamba LLMs have a tendency to gradually lose fine-grained visual details through intermediate layers, as shown in Fig. \ref{fig:compare}. This could be due to the inherent lack of 'positional embedding'-analogues within the Mamba LLM backbone, such that spatial and fine-grained information is more easily distorted inside the LLM. Consequently, we contend that the visual feature derived from the final layer of Mamba LLM alone proves inadequate in preserving fine visual intricacies. To enable the Mamba LLM to extract more sufficient visual details and prevent gradual feature loss from the image modality, we devise a \textbf{Multi-scale Feature Fusion} (MFF) module that \textit{hierarchically} integrates multiple intermediate features of the pretrained visual encoder for the visual alignment. Combining multi-level features that concern different levels of granularity enables the model to effectively capture intricate details at various scales, enhancing its ability to comprehend and interpret complex visual information with greater precision. Specifically, our multi-scale feature fusion module consists of I fusion blocks $\psi = \{\mathcal{B}_i\}^I$, where each $\mathcal{B}$ can be regarded as a tiny Mamba network consisting of a cross-attention module and a Mamba layer, which combines intermediate features in a pairwise fashion. For instance, given hidden visual features $\{\overline{X}_i, \overline{X}_j, \overline{X}_k\}$ from layers $i$, $j$, $k$ and the corresponding two-layer MFF $\psi = \{\mathcal{B}_i\}^2$, the final aggregated output of the features $\overline{X}_{v}$ is generated as follows:
\begin{equation}\overline{X}_{v}= \psi(\overline{X}_i, \overline{X}_j, \overline{X}_k)=\mathcal{B}_2(\mathcal{B}_1(\overline{X}_i, \overline{X}_j), \overline{X}_k)
\end{equation}
Where each block $\mathcal{B}$ is given by residually-connected Mamba and cross attention layers:
\begin{equation}\mathcal{B}(X,Y)=\hat{\mathcal{B}}(X,Y)+Mamba(\hat{\mathcal{B}}(X,Y)); \quad \hat{\mathcal{B}}(X,Y)=X+cross\_attn(X,Y)
\end{equation}
In practice, we utilize three intermediate layers along with the final output layer for producing the aggregated visual feature for pixel-wise alignment. Thus, the final pixel-wise alignment becomes:
\begin{equation}\mathcal{L}_{pixel}=\| f_{dec}(\overline{X}_v) - X_v \|^2_{2} 
\end{equation}
By introducing additional contribution of intermediate features in the pixel-wise alignment loss, we force the model to retain structural and fine-grained information in these layers and alleviate the gradual loss of visual features. This results in better-quality visual features, as evident in Fig. \ref{fig:compare}. Additionally, we note that the feature fusion and visual decoding stage only occurs during training where loss calculations are needed, and poses no additional computational overhead in inference.

% Although the
% vision encoder has a global reception field in all layers, it
% is verified that different transformer layers learn visual information at different scales [10], e.g., lower layers learn visual details. Thus, our vision aggregator makes fine-grained
% spatial-aware visual knowledge more likely to be learned
% based on visual grounding tasks. Specifically, our vision
% aggregator can be regarded as a tiny transformer-style network, consisting of two transformer layers for aggregating
% the hidden features from the vision encoder. Given the
% hidden features {Vi
% , Vj , Vk} from some middle layers in
% the vision encoder, the vision aggregation module uses two
% blocks to sequentially integrate the former two features with
% the last feature. Each block B is composed of self attention
% (Attn), cross attention (XAttn), and Feed-forward network
% (FFN) arranged in a sequential manner. Finally, the output
% features V¯ is generated as follows,
% V¯ = B2(B1(Vi
% ; Vj ); Vk), (4)
% B(X; Y ) = FFN(XAttn(Attn(X), Y )). (5)
% In practice, we use the middle layers {i = L − 1, j =
% 2L/3, k = L/3} in the vision encoder to produce the hidden features as the input to VA, where L is the number of
% layers in the vision encoder.

\section{Experiments}
In this section, we conduct extensive experiments from multiple perspectives to demonstrate the effectiveness of our method. First, we provide experiment settings including training data, training recipes, and evaluation benchmarks in Sec. \ref{sec:set}. Next, we provide experimental results from the aforementioned benchmarks and compare with contemporary MLLMs in Sec. \ref{sec:res}. EMMA surpasses other Mamba-based models on a majority of tasks and is highly competitive with transformer models of similar size. Then, we conduct a detailed inference speed comparison between EMMA and other Mamba and transformer-based models of similar size in Sec. \ref{sec:latent}. At last, we present ablation studies to investigate the effectiveness of our model design choices in Sec. \ref{sec:abl}.

% model along with the quantitative and qualitative analyses. Please refer to Appendix for
% implementation details and training details.

% Due to the recent release of MambaV2 \cite{dao2024transformers}, we include both MambaV1 and MambaV2 variants in 

\subsection{Experiment Settings}\label{sec:set}

\textbf{Training Data.} Following \cite{zhao2024cobra}, we train EMMA on a combination of datasets consisting of LLaVA-v1.5-mixed-665k \cite{liu2024visual}, LVIS-Instruct-4V \cite{wang2023see}, and LRV-Instruct \cite{liu2023mitigating}. These are conversational multi-modal and pure textual data that contains roughly 1.2 million images and dialogues. 

\textbf{Backbone Models.} We select cobra as our baseline model. Following \cite{zhao2024cobra}, we utilize SigLIP and DINOv2 as the visual encoder, where the final output is the concatenated feature of both models. The image decoder consists of a combination of 4 Mamba and linear layers. The MLP projector consists of stacked linear layers. We provide two versions of our model with MambaV1-2.8b \cite{gu2023Mamba} and MambaV2-2.7b \cite{dao2024transformers} backbones. 

\textbf{Training Recipes.} We directly finetune the Mamba LLM backbone, the multi-scale fusion module, the image decoder, and the MLP projector on the training data for two epochs, discarding the pretrain phase. The visual encoder is frozen at all times. We select a global batch size of 128 and a starting learning rate of 2e-5 with AdamW optimization. Our models are trained on eight 40G A100 GPUs with fully sharded data parallelism \cite{zhao2023pytorch}. 

\textbf{Evaluation Benchmarks.} We evaluate our model variants on four open-ended visual question-answer benchmarks: VQAv2 \cite{goyal2017making} and VizWiz \cite{gurari2018vizwiz} test general visual reasoning, GQA \cite{hudson2019gqa} validates spatial reasoning, and TextVQA \cite{singh2019towards} assesses reasoning around text. We also evaluate our models on nine comprehensive closed-set benchmarks: VSR \cite{liu2023visual} tests coarse object spatial relationships. POPE \cite{li2023evaluating} tests object hallucinations while HallusionBench \cite{guan2024hallusionbench} assesses object illusion and visual hallucinations. MMB \cite{liu2023mmbench} and MME \cite{fu2024mmecomprehensiveevaluationbenchmark} are both robust and holistic evaluations of MLLMs, SEED-IMG \cite{li2024seed} evaluates generative comprehension. AI2d \cite{hiippala2021ai2d} and ScienceQA \cite{saikh2022scienceqa} both evaluate science-related topics. CCBench is a benchmark on Chinese culture \cite{liu2023mmbench}. We use reproduced results from the Cobra codebase \cite{zhao2024cobra} and obtain the performance of other models in their respective papers. More details for experimental settings can be found in the Appendix. 

\subsection{Evaluation Results}\label{sec:res}
\begin{table*}[htb]
\centering
\label{tab2}
\resizebox{\textwidth}{!}{
\begin{tabular}{lcccccccccc}
\hline Model & LLM & Data Size & VQA$^{\text {v2 }}$ & GQA & VizWiz & VQA$^{T}$ & VSR & MME & MMB \\
\hline OpenFlamingo & MPT-7B & 2B & 52.7 & - & 27.5 & 33.6 &  - & - & -\\
BLIP-2 & Vicuna-13B & 129M & - & 41.0 & 19.6 & 42.5 & 50.9 &  1293.8 & -\\
MiniGPT-4 & Vicuna-7B & 5M & 32.2 & - & - & - & -  & 581.7 & -\\
EMU & LLaMA-13B & 3.4B & 62.0 & 46.0 & 38.3 & - &  - & - & -\\
EMU 2 & LLaMA-33B & 4.1B & 84.9 & 65.1 & 54.9 & 66.6 &  - & - & - \\
InstructBLIP & Vicuna-7B & 130M & - & 49.2 & 34.5 & 50.1 & 54.3 &  - & 36.0 \\
InstructBLIP & Vicuna-13B & 130M & - & 49.5 & 33.4 & 50.7 & 52.1 &  1212.8 & - \\
Shikra & Vicuna-13B & 6.1M & 77.4 & - & - & - & - &  - & 58.8 \\
IDEFICS & LLaMA-7B & 354M & 50.9 & - & 35.5 & 25.9 & - & - & 48.2\\
IDEFICS & LLaMA-75B & 354M & 60.0 & - & 36.0 & 30.9 &  - & - & 54.5\\
Qwen-VL & Qwen-7B & 1.5B & 78.2 & 59.3 & 35.2 & 63.8 &  - & - & 38.2\\
LLaVA v1.5 & Vicuna-7B & 1.2M & 78.5 & 62.0 & 50.0 & 58.2 & - & 1510.7 & 64.3 \\
Prism & LLaMA-7B & 1.2M & 81.0 & 65.3 & 52.8 & 59.7 & 59.6 &  - & - \\
ShareGPT4V & Vicuna-7B & 1.2M & 80.6 & 57.2 & - & - & -  & 1567.4 & 68.8 \\
MoE-LLaVA & StableLM-1.6B x 4 &  2.2M & 76.7 & 60.3 & 36.2 & 50.1 & - &  - & 59.4 \\
MoE-LLaVA & Phi2-2.7B x 4&  2.2M & 77.6 & 61.4 & 43.9 & 51.4 & - &  - & 65.5 \\
\hline LLaVA-Phi & Phi2-2.7B &  1.2M & 71.4 & - & 35.9 & 48.6 & - &  1335.1 & 59.8 \\
MobileVLM-3B &  MobileLLaMA-2.7B & 1.2M & - & 59.0 & - & 47.5 & - &  1288.9 & 59.6  \\
MobileVLM v2 & MobileLLaMA-2.7B & 3.6M & - & $\mathbf{61.1}$ & - & $\mathbf{57.5}$ & - &  1440.5 & 63.2\\
% TinyLLaVA & Phi2-2.7B & $\mathbf{79.9}$ & $\mathbf{62.0}$ & - & $\mathbf{59.1}$ & - & 86.4 \\
TinyLLaVA & Phi2-2.7B &  1.2M & $\mathbf{\underline{76.6}}$ & 60.3 & - & 51.4 & - & 1464.9 & $\mathbf{66.9}$ \\
\hline VL-Mamba & MambaV1-2.8B &  1.2M & $\mathbf{\underline{76.6}}$ & 56.2 & - & 48.9 & - &  1369.6 & 57.0 \\
Cobra & MambaV1-2.8B & 1.2M & 74.9 & 59.1 & 52.0 & 52.4 & 51.7  & 1294.3 & 51.5 \\
% Cobra+MambaV2^$*$ & MambaV2 LLM-2.7B & 75.71 & 58.2 & 52.34 & 55.8 & 52.9 & $\mathbf{87.3}$ \\
% ML Mamba & MambaV2 LLM-2.7B & 75.26 & $\mathbf{60.68}$ & 45.17& 52.2 & 51.5 & $\mathbf{88.3}$ \\
EMMA-V1 & MambaV1-2.8B & 1.2M & 76.3 & $\underline{60.5}$ & $52.1$ & $\underline{57.2}$ & $51.5$ & $\mathbf{\underline{1572.8}}$ & $53.2$ \\ 
EMMA-V2 & MambaV2-2.7B & 1.2M & 75.7 & 59.4 & $\mathbf{\underline{54.1}}$ & 56.2 & $\mathbf{\underline{52.7}}$ &  1454.9 & $\underline{60.8}$
\\
\hline
\end{tabular}
}
\caption{Experimental results on seven multi-modal benchmarks: VQA$^{\mathrm{v} 2}$, GQA, VizWiz, TextVQA, VSR, MME and MMB. We separate the models into three groups by horizontal lines: the first group consists of large MLLMs, the second group consists of small-scaled transformer MLLMs, and the third group consists of Mamba MLLMs. Our method surpasses other Mamba-based frameworks in all but the VQA$^{\mathrm{v} 2}$ task, where VL Mamba has a slight advantage of less than 1\%. Our method also achieves competitive results with other transformer-based methods of similar scale, while attaining a much smaller computational complexity, as detailed in Sec. \ref{sec:latent}.}
\label{tab_all}
\end{table*}

We compare our model with a variety of Mamba-based and transformer-based MLLMs. These include large-scaled MLLMs: OpenFlamingo \cite{awadalla2023openflamingo}, BLIP-2 \cite{li2023blip}, MiniGPT-4 \cite{zhu2023minigpt}, EMU \cite{sun2023emu}, EMU 2 \cite{sun2024generative}, InstructBLIP \cite{dai2023instruct}, Shikra \cite{chen2023shikra}, IDEFICS \cite{hua2024talk}, Qwen-VL \cite{bai2023qwen}, LLaVA \cite{liu2024visual}, Prism \cite{karamcheti2024prismatic}, ShareGPT4V \cite{chen2023sharegpt4v}, MoE-LLaVA \cite{lin2024moe}. We also include transformers with similar scales, which encompass LLaVA-Phi \cite{zhu2024llava}, MobileVLM \cite{chu2023mobilevlm}, MobileVLM V2 \cite{chu2024mobilevlm}, and TinyLLaVA \cite{zhou2024tinyllava}. Lastly, we also compare with current Mamba-based MLLMs, consisting of VL Mamba \cite{qiao2024vl} and Cobra \cite{zhao2024cobra}. We also provide the backbone LLM of each model, as well as the training data size for more fair comparison. 

\textbf{Comparison with similar-scaled Mamba MLLMs.}
Performance comparisons between EMMA and other Mamba-based MLLMs can be found in the bottom group of Tab. \ref{tab_all} as well as \ref{add_comp}. We underline the best-performing models in this category. All models utilize a similar amount of data and Mamba backbone. We surpass the performance of cobra, our baseline model, on every evaluated metric, demonstrating better generalizability as well as visual and spatial reasoning. The increase in performance is most apparent in MME, where there is a noticeable increase of $279$ in terms of the general perception and cognition abilities of the model. There is also a $5\%$ increase in the open-ended VQA task TextVQA, which reflects the attentiveness of fine-grained details of our model. The performance gain is least apparent in VSR due to it consisting solely of true/false answers, which are inherently coarse in nature and do not necessitate fine-grained details. Nevertheless, our model still improves upon the cobra baseline on these benchmarks with small gains. We also outperform VL Mamba on all but the VQAv2 benchmark, where our best model underperforms by 0.3. These results suggest that Mamba MLLMs benefit from better-quality visual features through structural and hierarchical alignment, resulting in overall gains across the majority of metrics.

% https://huggingface.co/spaces/opencompass/open_vlm_leaderboard
\begin{table*}[htb]
\centering
\begin{tabular}{lcccccc}
\hline Model & SEED & ScienceQA & AI2D & CCBench \\
\hline 
Cobra  &  61.5 & 63.0 & 48.6 & 11.6 \\
% VL Mamba &  - & - & -\\
EMMA-V1 (Ours)  & 62.9 & $\mathbf{66.3}$ & $\mathbf{50.5}$ & 16.7 \\ %34.3-0.1 if add '\nAnswer the question using a single word or phrase 
EMMA-V2 (Ours) & $\mathbf{64.3}$ & 65.0 & 49.1 & $\mathbf{29.21}$ \\
\hline
\end{tabular}
\caption{More performance comparisons across MLLMs with similar sizes. Since none of the other 3b-scale MLLMs report performance on these benchmarks, we report the results for cobra and ours.}
\label{add_comp}
\end{table*}

\textbf{Comparison with similar-scaled transformer MLLMs.}
Performance comparisons between EMMA and other transformer-based MLLMs can be found in the middle group of Tab. \ref{tab_all}. We bold-font the best-performing models for all similar-scaled MLLMs, Mamba and transformer-based. Note that although these models have similar parameter sizes to ours, their backbone LLM was trained on 1.3T (MobileLLaMA) and 1.4T (Phi-2-2.7B) textural tokens prior to the MLLM pretrain and finetune stage. On the other hand, MambaV1-2.8b was trained on the SlimPj \cite{shen2023slimpajama} and MambaV2-2.7b on the Pile \cite{gao2020pile}, with only 627B and 300B tokens respectively. Furthermore, MobileVLM V2 trains on nearly three times as much data as ours prior to evaluation. Other networks such as TinyLLaVA train on the more diverse ShareGPT4V dataset, which greatly improves model performance. Nevertheless, our model achieves the best VizWiz, VSR, and MME scores across all models of similar scale. Our model is also competitive on other metrics, many of which are only \(\sim1\) below the best-performing model.

\textbf{Comparison with large-scaled transformer MLLMs.} Lastly, we also compare EMMA with other large-scaled MLLMs in the upper group of Tab. \ref{tab_all}. Although there is a significant gap in model size and training data, EMMA still achieves competitive performance in some benchmarks. We also highlight EMU and EMU2, which is a transformer MLLM that also jointly generates visual and textual tokens in an autoregressive fashion. However, both EMU and EMU2 require a sizable stable diffusion decoder to translate visual tokens into image space, along with additional training stages to optimize their latent space. These transformer-based models also demand a substantial amount of data and possess significantly larger parameter counts compared to our model. Regardless, our model consistently outperforms EMU across all benchmarks while utilizing nearly one-fourth of its parameters. EMMA also achieves the best MME across all competing models. EMMA is second in terms of VizWiz, only 0.8 behind the best-performing model EMU2, which is nearly ten times bigger than our model. These results demonstrate the effectiveness of EMMA and show the potential of Mamba-based MLLMs as an alternative to transformer-based models.

\begin{table*}[htb]
\centering
\resizebox{\textwidth}{!}{
\begin{tabular}{lccc|cccccc}
\hline & EMMA-V1 & EMMA-V2 & Cobra & GPT4V & LLaVA-1.5 & Claude3 & BLIP2-T5 & Qwen-VL & MiniGPT5\\
\hline POPE & \textbf{88.0} & 87.3 & 87.2 & - & 85.9 & - & - & - & - \\
% VL Mamba & Mamba LLM-2.8B & - & - \\ 
HBench & 51.0 & 47.5 & 41.4 & \textbf{65.28} & 47.1 & 56.9 & 48.1 & 39.2 & 40.3\\ 
\hline
\end{tabular}
}
\caption{Performance comparisons for hallucination-related benchmarks on a selection of MLLMs. HBench denotes HallusionBench. Our model achieves the best performance in POPE and is closely competitive with large-sized models. }
\label{tab_hal}
\end{table*}

\textbf{Analysis for Hallucination.}
We analyze the degree of hallucination in our models and report corresponding performances on the POPE and HallusionBench benchmark in Tab. \ref{tab_hal}. POPE concerns object hallucination which refers to generated contents inconsistent with ground-truth objects in the input, which reflect hallucination leaning towards the language modality. HallusionBench focuses on diagnosing both the visual illusion and knowledge hallucination of MLLMs. Visual illusion refers to the misinterpretation of accurate visual information, and knowledge hallucination denotes perceptions formed without relevant visual input. Thus, HallusionBench reveals hallucinations more on the effective extraction of information in the visual modality. Our model achieves the highest score on POPE, demonstrating the superior NLP capabilities of the Mamba LLM backbone. We also achieve significant gain (51.0 vs 41.4) in HallusionBench from our baseline, demonstrating the effectiveness of higher-quality visual features in reducing visual hallucinations. Our model is also competitive in HallusionBench with LLaVA-1.5 and BLIP2-T5, a 7B and 12.4B model respectively. 

\subsection{Model Latency}\label{sec:latent}
We evaluate model latency between EMMA and other similar-sized MLLMs, which include cobra, TinyLLaVA, and MobileVLM V2 in Tab. \ref{tab_latent}. While VL Mamba is not open-sourced yet, we assume the latency of VL Mamba to mirror that of Cobra, as both models leverage the MambaV1-2.8b backbone architecture. All models are given the same sample image and the exact textural prompt of "Describe this image in detail". Each model is then forced to generate 256 tokens in response to this prompt for a total of 200 times, and the overall time taken is recorded as $T_{overall}$. Finally, we calculate the average time taken for each model to generate 256 tokens ($T_{avg}$) as well as the number of tokens generated per second ($N_{avg}$) as follows:
\begin{equation}
T_{avg}=T_{overall}/200; \quad N_{avg}=256/T_{avg}
\end{equation}
All evaluations are conducted on a single 40G A100 GPU. EMMA has a significant time advantage over transformer-based MLLMs and a non-trivial edge over other Mamba-based MLLMs. Mamba models, given their better theoretical guarantees, are three to five times faster than current state-of-the-art transformer-based models of similar sizes. Our model achieves even better runtime than cobra due to more efficient processing in the MambaV2 LLM backbone. 

\begin{table*}[htb]
\centering
\begin{tabular}{llcc}
\hline Model & LLM Backbone & Tokens per second & Average Time\\
\hline EMMA (Ours) & MambaV2 LLM-2.7B & 149.96 & 1.71 \\
% VL Mamba & Mamba LLM-2.8B & - & - \\ 
Cobra & Mamba LLM-2.8B & 138.95 & 1.84 \\ 
TinyLLaVA & Phi2-2.7B & 41.46 & 6.17 \\
MobileVLM V2 & MobileLLaMA-2.7B & 39.36 & 6.50 \\ 
\hline
\end{tabular}
\caption{Latency table for similar-scaled MLLMs, both Mamba \& transformer architectures.}
\label{tab_latent}
\end{table*}

\subsection{Ablation studies}\label{sec:abl}
In this section, we present ablation studies for the design choices of our model in Tab. \ref{ab_table}.

\textbf{Component-wise Ablation.} We first explore the impact of the two proposed components on the overall performance of our model. By removing the multi-scale feature fusion module, we observe a noticeable reduction in MME, as well as slight degradation in VQAv2, GQA, POPE, and HallusionBench. This shows that the multi-scale feature fusion module plays a crucial role in enhancing performance across multiple tasks, by preventing visual information loss in the LLM, thereby enhancing the overall quality of visual features. We then remove the pixel-wise alignment loss, which is equivalent to training the plain cobra model. We observe a substantial decline in TextVQA and HallusionBench, along with marginal decreases in GQA, VQA-V2, and MMB. Notably, TextVQA involves open-set VQA, relying heavily on intricate visual details, whereas HallusionBench addresses visual hallucinations. This outcome underscores the impact of our pixel-wise alignment loss in improving the quality of visual representations, leading to enhanced processing of fine visual details and a reduction in visual hallucinations.

\begin{table*}[htb]
\centering
\resizebox{\textwidth}{!}{
\begin{tabular}{llcccccccc}
\hline Model & VQA$^{\text{v2}}$ & GQA & VizWiz & VQA$^{T}$ & VSR & POPE & MME & MMB & HBench \\

% \hline MMIT (Ours) & 75.65 & $\mathbf{59.37}$ & 54.1 & $\mathbf{56.2}$ & $\mathbf{52.7}$ & 87.3 \\
\hline EMMA (Ours) & $\mathbf{76.25}$ & $\mathbf{60.5}$ & 52.1 & $\mathbf{57.2}$ & 51.5 & $\mathbf{88.0}$ & $\mathbf{1572.8}$ & $\mathbf{53.2}$ & $\mathbf{51.0}$\\

% \hline $-$multi-scale feature & 75.62 & 59.25 & $\mathbf{54.55}$ & 55.2 & 51.5 & 87.3 & 1294.1 & 53.1 \\ %56.2
% $-$pixel-alignment loss & 75.71 & 58.2 & 52.34 & 55.8 & 52.5 & 87.3 & 1294.3 & 51.5 \\ %52.9

\hline $-$MFF & 75.9 & 59.3 & 52.0 & 57.0 & 51.5 & 87.1 & 1294.1 & 53.1 & 50.7 \\ %56.2
$-$PAL & 74.9 & 59.1 & 52.0 & 52.4 & $\mathbf{51.7}$ & 87.2 & 1294.3 & 51.5 & 41.4 \\ %52.9
\hline $+$CSM & 75.8 & 59.0 & $\mathbf{54.2}$ & 55.6 & 51.5 & 87.6 & 1420.5 & 52.7 & 50.5 \\ % last two no 
$+$AVF & 52.8 & 44.8 & 45.0 & 41.1 & 51.6 & 69.7 & 984.8  & 25.0 & 47.2 \\ %52.9 last two no 

\hline
\end{tabular}
}
\caption{Ablation table for our model. Without loss of generality, we utilize the MambaV1 version of our model for these experiments. HBench denotes HallusionBench. MFF denotes the multi-scale feature fusion module, PAL denotes the pixel-alignment loss, CSM denotes the cross-scan mechanism, and AVF denotes align visual features.}
\label{ab_table}
\end{table*}

\textbf{Preserving Structural Information in MambaLLM.} We note that many works \cite{qiao2024vl, liu2024vMamba} utilize a visual selective scan mechanism to incorporate structural information and enhance the quality of visual features. Consequently, we opt to integrate this mechanism on top of our method (before the projection MLP) to assess its impact on the quality of visual representations and model performance (+CSM). The results demonstrate a marginal change in performance, suggesting that our structural constraint framework might already suffice in generating high-quality structural visual features. The introduction of the cross-scan module may appear to be unnecessary.

\textbf{Pixel vs. Feature Alignment.} Given that the Mamba LLM receives visual features from the vision encoder, we also explore directly aligning these processed features instead of the raw image pixels (+AVF). However, the resulting performance experiences a notable decline across various metrics. We attribute this degradation to the robust structural information inherent in pixel-level images, contrasting with the extracted features from the visual encoder that may lack such details. However, despite the severe performance degradation, the model achieves a high hallusionBench score which implies the efficacy of supervising visual features in mitigating visual hallucinations.

% 19 is single-stage
% 17 is cross-scan
\section{Conclusion}
In this study, we introduce EMMA, a novel approach designed to rectify the imbalanced quality between visual and textual latents within the MLLM framework, which adversely impacts cross-modal alignment and leads to suboptimal performance. By expanding the autoregressive generation of text tokens to operate on image-wise patches, we leverage this process as visual supervision for the visual features within the Mamba LLM. Subsequently, we introduce a pixel-wise alignment loss to align the generated visual features with the original image, thereby preserving crucial structural information. On the other hand, the absence of structural constraints in the Mamba LLM leads to a gradual loss of fine-grained details in the visual features. To address this issue and counteract the gradual information loss, we propose a multi-scale feature fusion module that hierarchically integrates multiple intermediate LLM features, effectively preserving fine-grained information across these layers. Experimental results showcase that EMMA not only significantly enhances performance across diverse benchmarks but also exhibits reduced levels of hallucination. The MambaV2 iteration of our model further reduces model latency compared to current Mamba MLLMs and achieves nearly four times the speed of similarly scaled transformer models during inference. We hope that this paper opens up new ways for enhancing the visual latent quality of Mamba-based MLLMs, presenting a more efficient and performance-comparable alternative to transformer models, especially in scenarios demanding rapid responses, such as autonomous driving and robotics control.

\bibliography{iclr2025_conference}
\bibliographystyle{iclr2025_conference}

\newpage
\appendix
\section{Appendix}

\subsection{Mamba Preliminaries}
\def\figMamba#1{
\begin{figure*}[#1]
    \centering
    \includegraphics[width=0.65\linewidth]{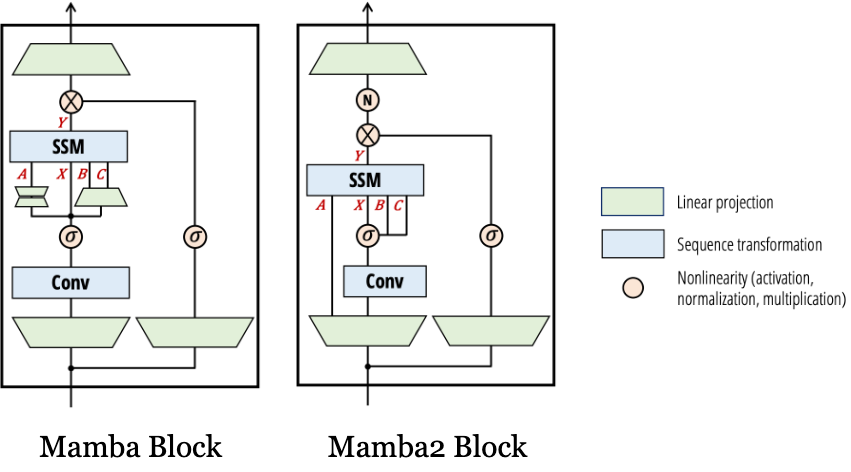}
    \caption{Comparison between Mamba and Mamba2 blocks \cite{dao2024transformers}.}
    \label{fig:Mamba}
\end{figure*}
}

Our MLLM is composed of Mamba \cite{gu2023Mamba} blocks, which are a form of structured state-space sequence models (S4) \cite{gu2021efficiently} with an input-dependent selective scanning function. We provide preliminaries regarding S4 models and the Mamba architecture in this section. Consider a sequential input $x \in \mathcal{R}^n$, the S4 model is defined by a latent state $h(t) \in \mathcal{R}^n$ with four parameters ($\Delta$, A, B, C) that maps $x$ to another sequence $y \in \mathcal{R}^n$, given by the following equation:

\begin{equation}
\begin{aligned}
h^{\prime}(t) &= A h(t) + B x(t) \\
y(t) &= C h(t)
\end{aligned}
\end{equation}

Where ($\Delta$, A, B) are continuous parameters that require discretization rules to transform into discrete parameters that enable parameter computations, given by:

$$
\bar{A}=\exp (\Delta A) \quad \bar{B}=(\Delta A)^{-1}(\exp (\Delta A)-I) \cdot \Delta B
$$

Then, the discretized form of S4 can be computed as follows: 

\begin{equation}
\begin{aligned}
h_t &=\bar{A} h_{t-1}+\bar{B} x_t \\
y_t &=C h_t
\end{aligned}
\end{equation}

This endows S4 models with beneficial properties such as resolution invariance and automatic normalization, and can be connected to gating functions in RNN. It also allows these models to be computed as equivalently a linear recurrence or a global convolution:

\begin{equation}
\bar{K}=\left(C \bar{B}, C \overline{A B}, \ldots, C \bar{A}^k \bar{B}, \ldots\right) \Leftrightarrow  y=x * \overline{\boldsymbol{K}}
\end{equation}

\figMamba{!t}
Building upon the S4 framework, Mamba introduces a selective-scan mechanism (SSM) that is input-dependent. The main difference between S4 and Mamba is simply making the parameters ($\Delta$, B, C) as functions of the input sequence $x$, which allows the model to selectively remember or forget information depending on the current data in sequence. Due to the efficient computation for linear recurrence, Mamba is efficiently implemented and achieves significantly lower throughput time compared to transformers. Even without self-attention and MLP layers, Mamba achieves state-of-the-art performance in language, audio, and genomics tasks. Recently, there has also been ongoing research to further improve the architecture and efficiency of the Mamba layer. \cite{dao2024transformers} draws the association between linear recurrence and the attention mechanism, and proposes Mamba2 with slight modifications to the Mamba architecture. Specifically, they remove the sequential linear projections and produce the SSM parameters A, B, C at the beginning of the block instead of functions of the input $x$. They also introduce an additional normalization layer for better optimization stability. Mamba2 achieves even lower throughput than Mamba and outperforms it on language benchmarks with slightly fewer parameters. A comparison between Mamba and Mamba2 architecture is shown in Fig. \ref{fig:Mamba}.

% Difference between Mamba vs. transformers. 

% (1a) $\quad h_t=\bar{A} h_{t-1}+\bar{B} x_t$
% (2a) $\bar{K}=\left(C \bar{B}, C \overline{A B}, \ldots, C \bar{A}^k \bar{B}, \ldots\right)$
% (1b) $\quad y_t=C h_t$
% (2b) $y=x * \overline{\boldsymbol{K}}$

\subsection{Additional Training Details}
In this section, we provide additional training details for our model. Given that Cobra serves as our base model, the chosen values closely mirror those employed within the Cobra framework. Model configurations and hyperparameter selection can be found in Tab. \ref{hyp_table}.

\begin{table*}[htb]
\centering
\resizebox{0.6\textwidth}{!}{
\begin{tabular}{lc}
\hline Configuration &  \\
\hline Vision Encoder & DINOv2 + SigLIP ViT-SO \\
Backbone LLM & MambaV1-2.8b / MambaV2-2.7b \\ 
Projector & GeLU-MLP \\ 
Img Resolution & 384 x 384 \\
Img Tokens & 729 \\
Global Batch Size & 128 \\
Training Steps & 19K \\
Optimizer & AdamW \\
LR scheduler & Cosine Decay \\
Initial LR & 2e-5 \\
Weight Decay & 0.1 \\
Warmup Ratio & 0.03 \\
Number of Epochs & 2 \\
\hline
\end{tabular}
}
\caption{Detailed training configurations used for EMMA.}
\label{hyp_table}
\end{table*}

\textbf{Model Architecture and Hyperparameters.} We directly use off-the-shelf DINOv2 and SigLIP models as our vision encoder. The GeLU-MLP is simply a three-layer MLP with the GeLU activation function in between. For the MambaV1 model, we use the Mamba-2.8b-Zephyr version of it. The Mamba2 model is directly taken from the official github page, which is trained on the 300B Pile dataset. We select the version without attention. 

\textbf{Training Data.} We provide more details regarding the training data for our model, which consists of three subsets: 

1. LLaVA-Mixed-665K is a mixture of instruction-following multimodal data that contains a variety of VQA, OCR, region-level VQA, visual conversation, and language conversation data.

2. LVIS-INSTRUCT4V contains 220K pairs of image-text instructions that are generated by GPT-4v. These instructions are visually aligned and context-aware to improve cross-modal alignment.

3. LRV-Instruction covers 16 vision-and-language tasks with open-ended instructions and answers. It includes both positive and negative instructions for more robust visual instruction tuning.

\subsection{Additional Evaluation Details}

In this section, we offer a concise summary of the key aspects that each benchmark emphasizes when assessing the model in order to give an overview of the strengths of our model.

\textbf{VQAv2} \cite{goyal2017making} consists of 265K image-question-answer tuples sourced from the COCO dataset \cite{lin2014microsoft}. The evaluation of models in the VQAv2 test set centers on visual recognition, visual grounding, spatial reasoning, and language comprehension.

\textbf{GQA} \cite{hudson2019gqa} consists of 22M questions of daily images, where each image is associated with a scene graph of the image's objects, attributes, and relations, generated from the Visual Genome dataset \cite{krishna2017visual}. The GQA test set rigorously assesses the models' proficiency in visual and compositional reasoning.

\textbf{VizWiz} \cite{gurari2018vizwiz} is derived from a natural visual question-answering scenario, where visually impaired individuals captured images and verbally posed questions about them. Each image is then accompanied by 10 answers from crowd-sourcing. This benchmark tests predicting answers to visual questions and determining whether a visual question is unanswerable.

\textbf{TextVQA} \cite{singh2019towards} is a multimodal dataset consisting of 45K questions on 28K images. This benchmark not only tests recognizing textual information in the given images but also reasoning over them.

\textbf{VSR} \cite{liu2023visual} comprises caption-image pairs with true/false labels. Each caption delineates the spatial relationship between two distinct objects within the image. In this task, the model is tasked with determining whether the caption accurately describes the image or not.

\textbf{MME} \cite{fu2024mmecomprehensiveevaluationbenchmark} stands as a comprehensive benchmark for evaluating MLLMs, measuring both perception and cognition across a total of 14 subtasks. The perception portion includes the recognition of coarse-grained and fine-grained objects. The cognition part includes commonsense reasoning, numerical calculation, text translation, and code reasoning. When evaluating our model, we add the scores from perception and cognition tasks as our final score.

\textbf{MMB} \cite{liu2023mmbench} is also a comprehensive benchmark for evaluating MLLMs. It surpasses existing similar benchmarks in terms of the number and variety of evaluation questions and abilities. Answers are recorded through multiple-choice questions in both English and Chinese versions, enabling performance comparisons under a bilingual context. 

\textbf{SEED} \cite{li2024seed} encompasses 19K multiple-choice questions with precise human annotations. It covers 12 evaluation dimensions, encompassing the understanding of both image and video modalities. We only select the image subset of the SEED benchmark to evaluate our models.

\textbf{ScienceQA} \cite{saikh2022scienceqa} features scientific questions and answers gleaned from academic sources. Models are evaluated on their ability to reason with scientific knowledge using questions, answer choices, and relevant contexts.

\textbf{AI2D} \cite{hiippala2021ai2d} is a dataset comprising more than 5K grade school science diagrams with over 15K corresponding multiple-choice questions. This benchmark, like ScienceQA, also tests knowledge of MLLMs on science-related topics.

\textbf{CCBench} \cite{liu2023mmbench} is a part of the MMBench project that evaluates the models on the domain of Chinese Culture.

\textbf{POPE} \cite{li2023evaluating} targets the evaluation of object hallucination in MLLMs. Test samples include both positive and negative objects (non-existent objects), requiring models to accurately recognize positives and identify negatives. This evaluation benchmark measures hallucination more on the textual modality. 

\textbf{HallusionBench} \cite{guan2024hallusionbench} focuses on diagnosing both the visual illusion and knowledge hallucination of MLLMs. Visual illusion refers to the misinterpretation of accurate visual information, and knowledge hallucination denotes perceptions formed without relevant visual input. This benchmark ties more closely with testing hallucination on the visual modality.

\subsection{Additional Visualizations}

We also present additional visualizations for intermediate features in Mamba MLLMs to demonstrate the effectiveness of our method. As shown in \ref{fig:compare1}.

\def\figCompareo#1{
\begin{figure*}[#1]
    \includegraphics[width=\linewidth]{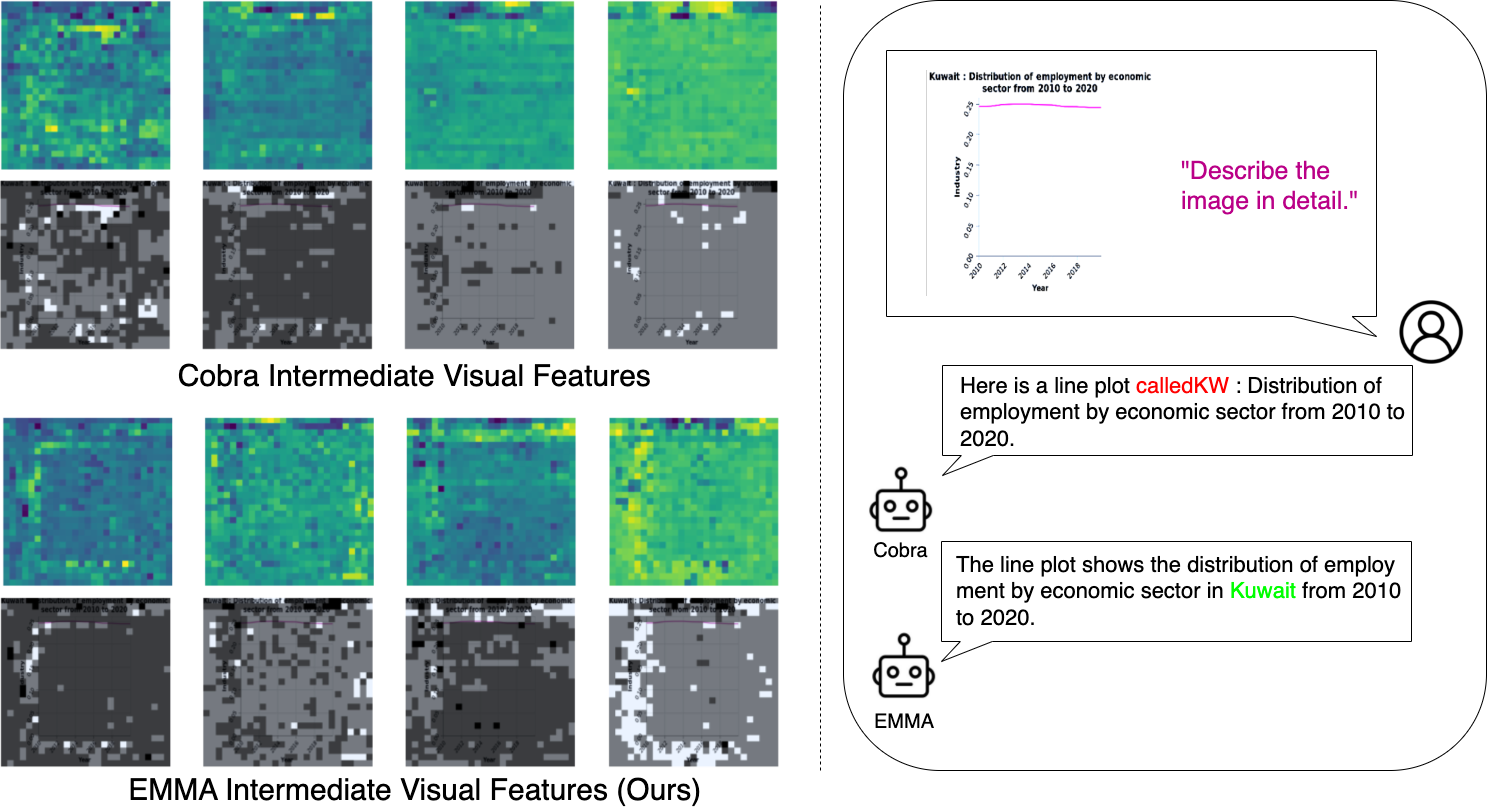}
    \label{fig:compare1}
    \caption{As shown, our method better highlights important characteristics of the image, such as the captions and horizontal and vertical axes. Cobra, on the other hand, gradually loses its visual activations on these characteristics, resulting in an incorrect reference to the plot as "calledKW".}
\end{figure*}
}

\figCompareo{}

\def\figComparer#1{
\begin{figure*}[#1]
    \includegraphics[width=\linewidth]{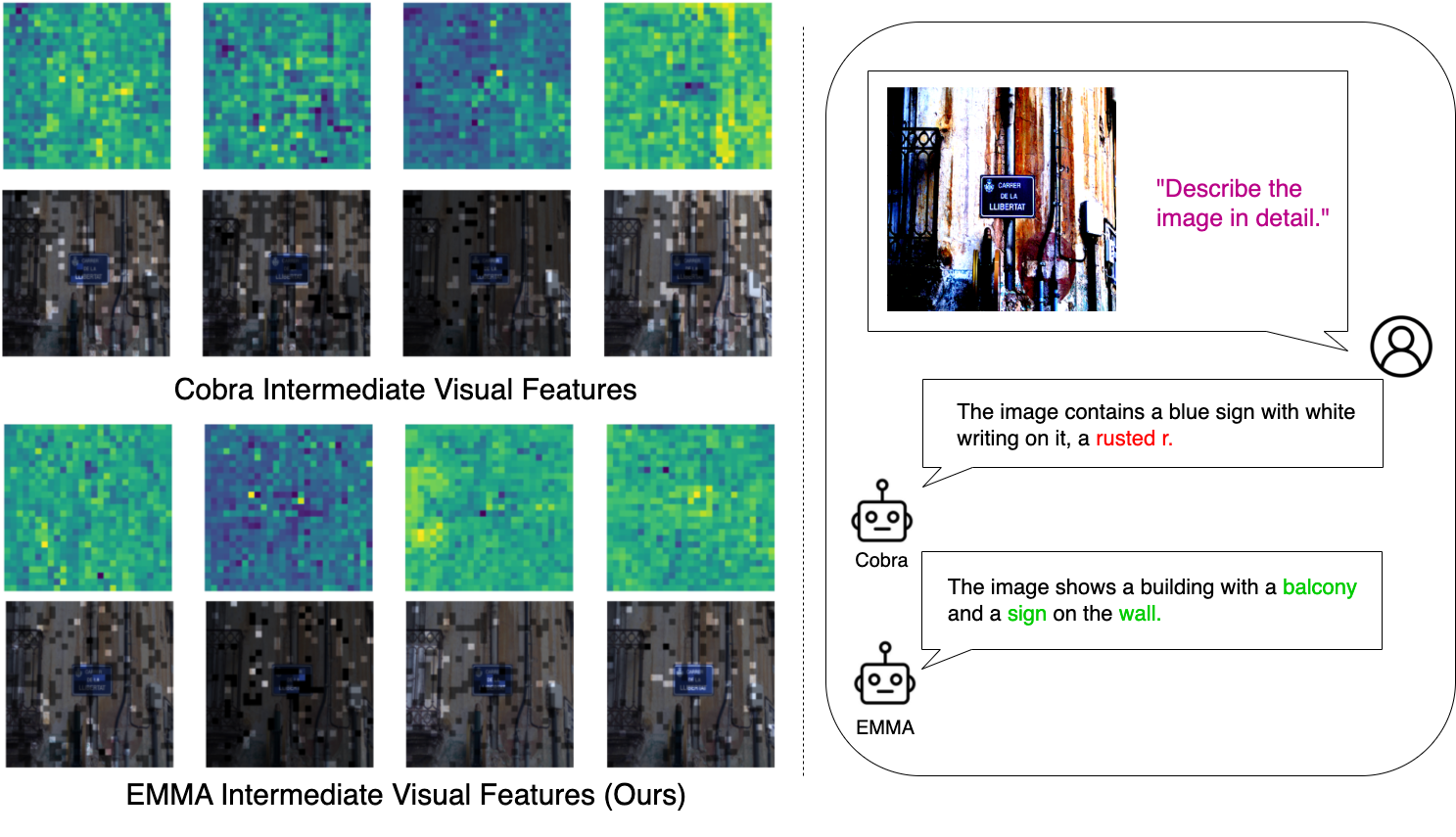}
    \label{fig:compare2}
    \caption{Similar to the previous example, EMMA is able to retain focus on important visual details even during later intermediate layers, resulting in better sensitivity to fine-grained visual details and less visual hallucination. }
\end{figure*}
}

\figComparer{}

% additional experimental details

% actual model implementation as given by cobra
% dataset in-depth explanation, ie mme, mmb etc.
% Mamba preliminary

\end{document}